\documentclass[sigconf,nonacm]{aamas}
\setcopyright{none}

\acmConference[arXiv]{arXiv preprint}{2025}{}

\renewcommand\footnotetextcopyrightpermission[1]{}

\usepackage{fancyhdr}

\fancypagestyle{firstpagestyle}{%
  \fancyhf{}%
  \fancyfoot[C]{ArXiv preprint. Under review.}%
}

\fancypagestyle{standardpagestyle}{%
  \fancyhf{}%
  \fancyfoot[C]{ArXiv preprint. Under review.}%
}

\usepackage{latexsym}
\usepackage{amsmath}
\usepackage{amsthm}
\usepackage{booktabs}
\usepackage{graphicx}
\usepackage{color}
\usepackage{algorithm}
\usepackage{algorithmic}
\usepackage{subcaption}
\usepackage{fancyhdr}
\usepackage{hyperref}

\newtheorem{theorem}{Theorem}
\newtheorem{lemma}[theorem]{Lemma}
\newtheorem{corollary}[theorem]{Corollary}

\newtheorem{definition}{Definition}
\newtheorem{remark}{Remark}
\newtheorem{assumption}{Assumption}

\usepackage{balance} %

\title{Beyond Predictions: A Participatory Framework for Multi-Stakeholder Decision-Making}\thanks{A preliminary version of this paper was accepted to MODeM$@$ECAI'25.\\Corresponding authors' email vineis@diag.uniroma1.it.\\
© 2025 Copyright held by the authors. ArXiv preprint.}

\author{Vittoria Vineis}
\affiliation{
  \institution{Sapienza University of Rome}
  \country{Italy}}

\author{Giuseppe Perelli}
\affiliation{
	\institution{Sapienza University of Rome}
	\country{Italy}}

\author{Gabriele Tolomei}
\affiliation{
	\institution{Sapienza University of Rome}
	\country{Italy}}

\begin{abstract}Conventional automated decision-support systems often prioritize predictive accuracy, overlooking the complexities of real-world settings where stakeholders’ preferences may diverge or conflict. This can lead to outcomes that disadvantage vulnerable groups and erode trust in algorithmic processes. Participatory AI approaches aim to address these issues but remain largely context-specific, limiting their broader applicability and scalability. To address these gaps, we propose a participatory framework that reframes decision-making as a multi-stakeholder learning and optimization problem. Our modular, model-agnostic approach builds on the standard machine learning training pipeline to fine-tune user-provided prediction models and evaluate decision strategies, including compromise functions that mediate stakeholder trade-offs. A synthetic scoring mechanism aggregates user-defined preferences across multiple metrics, ranking strategies and selecting an optimal decision-maker to generate actionable recommendations that jointly optimize performance, fairness, and domain-specific goals. Empirical validation on two high-stakes case studies demonstrates the versatility of the framework and its promise as a more accountable, context-aware alternative to prediction-centric pipelines for socially impactful deployments.
\end{abstract}

\keywords{ Participatory AI, Multi-stakeholder decision-making, Accountable socio-technical systems}

\newcommand{\BibTeX}{\rm B\kern-.05em{\sc i\kern-.025em b}\kern-.08em\TeX}

\begin{document}


\pagestyle{fancy}
\fancyhead{}


\maketitle 


\section{Introduction}
\label{sec:intro}
Advances in artificial intelligence (AI) and machine learning (ML) have boosted the development of \emph{automated decision-making (ADM) systems} in high-stakes domains such as healthcare, finance, and public policy \citep{chiusi2020automating}. However, the predominant \emph{top-down} design paradigm and the traditional optimization-oriented setup for these systems are increasingly under scrutiny, particularly due to their tendency to overlook critical stakeholder perspectives and broader societal impacts \citep{gerdon2022social}. In many cases, algorithmic solutions emphasize predictive accuracy above all else, neglecting or underperforming in aspects such as fairness \citep{petersen2023assessing}, transparency \citep{grimmelikhuijsen2023explaining}, and the consideration of heterogeneous and potentially conflicting interests \citep{laufer2023optimization}.  This narrow focus often leads to \emph{impact-blind} recommendations that risk perpetuating historical biases and exacerbating inequities \citep{barocas2016big}, highlighting the urgent need for more trustworthy, accountable systems \citep{lepri2018fair, floridi2021ethics, council2024regulation}. For example, in healthcare, predictive models for diagnosis or resource allocation may fail to account for disparities in service access or patient demographics, disproportionately harming marginalized communities \citep{obermeyer2019dissecting}. In finance, credit scoring models often embed biases from historical data, excluding individuals from economic opportunities \citep{hardt2016equality} and ignoring broader socioeconomic effects.

Recent efforts to mitigate these issues often involve post-hoc fairness adjustments or the adoption of generalized fairness metrics (e.g., \cite{mehrabi2021survey, pessach2022review}). While these approaches represent steps toward more equitable AI, they often fall short of addressing the nuanced, context-dependent priorities of real-world stakeholders. In particular, many current solutions remain \emph{single-perspective}\footnote{With this terms we refer to  systems that are not
multi-stakeholder aware. In this work, use the terms “stakeholder” and “actor” largely interchangeably,
the former rooted in social science, the latter in formal modeling
terminology.} in focus, thereby missing the opportunity to model and reconcile multiple, potentially conflicting preferences in a transparent manner. 
In this context, recent developments in \emph{participatory approaches to AI design} offer promising avenues, but often lack versatility across use cases and scalability.
 
To address these gaps, we propose a novel, modular, and flexible multi-stakeholder decision-making framework that reinterprets both the offline training and online deployment phases of standard predictive model–based ADM systems as a \emph{multi-actor participatory learning and decision-making process}. To the best of our knowledge, our framework is the first to operationalize participatory decision-making as an end-to-end, theoretically grounded, and adversarially robust learning process, bridging normative reasoning and computational implementation in multi-actor ADM.
Overall, our framework makes four main contributions: 

(1) \textbf{Theoretical unification} – It establishes a principled bridge across reward-based learning, game theory, welfare economics, computational social choice, and multi-objective optimization, providing a coherent and general theoretical foundation for participatory AI and its practical applications.

(2) \textbf{Domain- and model-agnostic modularity} – It supports scalable deployment across diverse high-stakes contexts through the dynamic incorporation of stakeholder preferences, without committing to fixed metrics, model architectures, or decision criteria. It remains fully compatible with existing predictive pipelines, enabling seamless integration into real-world ADM systems.

(3) \textbf{Structural accountability} – It offers explicit mappings between stakeholder preferences and collective outcomes, enabling both technical and non-technical actors to trace how individual priorities influence final decisions. This design promotes transparency, interpretability, and normative accountability in socio-technical environments where conventional systems remain opaque.

(4) \textbf{Robustness certificate schema} - It formalizes a general certificate schema that provides first-order lower bounds on the degradation of composite scores under coalition-bounded perturbations, parameterized by rule-specific smoothness and curvature constants. The schema generalizes across compromise operators, and we derive its structure and temperature-optimality analytically.

The remainder of the paper formalizes the proposed framework and validates it empirically on two high-stakes domains.  Comprehensive proofs, further details on the empirical validation are available in the Technical Appendix, while the full codebase is available at the following link: \\\href{https://anonymous.4open.science/r/participatory_training-502B/README.md}{https://anonymous.4open.science/r/participatory\_training-502B}

\section{Related Work}

\paragraph{Fairness and accountability in ADM} The pursuit of \emph{fairness and accountability} in ADM systems has attracted growing attention across both research and policy communities, reflecting their increasing deployment in high-stakes domains~\citep{chiusi2020automating} and associated societal impacts~\citep{araujo2020ai}. A large body of work has investigated pre-, in-, and post-processing techniques to mitigate bias in data and modeling pipelines (for comprehensive reviews, see~\citep{mehrabi2021survey, pessach2022review}).
Yet, evidence consistently shows that \emph{algorithmic fairness interventions alone do not guarantee fairness in practice}~\citep{goel2021importance, jeong2022fairness}. This is also due to the fact that \emph{fairness is tightly interdependent} with other system properties—most notably interpretability—whose joint operationalization remains challenging~\citep{dodge2019explaining, schoeffer2022relationship, ramachandranpillai2023fairxai, jain2020biased}. Moreover, fairness is inherently \emph{context-dependent}: cultural, institutional, and social norms shape how fairness principles are interpreted and applied, complicating efforts to design frameworks that extend beyond predefined universal metrics~\citep{selbst2019fairness, arif2022towards}. Even though well known  “impossibility theorems” \citep{kleinberg2016inherent,chouldechova2017fair} have been partially questioned in the fairness literature \citep{green2022escaping, hsu2022pushing, bell2023possibility}, they nonetheless highlight the fundamental tensions inherent in reconciling multiple, potentially incompatible fairness principles. Additionally, fairness-aware methods, while often effective at reducing specific statistical disparities, tend to overlook the systemic inequities and governance structures that mediate algorithmic outcomes. This limitation underscores the need for \emph{context-aware, multi-actor approaches} that integrate diverse stakeholder perspectives and explicitly connect fairness objectives with broader accountability mechanisms~\citep{gerdes2022participatory}.\\
\emph{Addressing these gaps, our framework does not enforce predefined fairness criteria, yet offers an accountable mechanism to transparently evaluate fairness and welfare properties as they emerge from the aggregation of heterogeneous stakeholder preferences.}

\paragraph{Game-theoretic and social choice foundations.}
The challenge of reconciling heterogeneous stakeholder preferences has deep roots in social choice theory and cooperative game theory. Classical impossibility results—most notably Arrow’s theorem~\citep{arrow1950difficulty,arrow1964social} and the Gibbard–Satterthwaite theorem~\citep{gibbard1973manipulation,satterthwaite1975strategy}—demonstrated that, under mild conditions, no aggregation rule can simultaneously satisfy Pareto efficiency, independence of irrelevant alternatives, and non-dictatorship, nor achieve collective fairness while remaining strategy-proof. These foundational results formalize the inherent trade-offs among efficiency, fairness, and strategic robustness that characterize collective decision-making.

Building on these insights, cooperative and welfare-oriented branches of social choice theory developed axiomatic bargaining frameworks—such as Nash bargaining~\citep{nash1950bargaining}, Kalai–Smorodinsky~\citep{kalai1975other}, and proportional fairness~\citep{Kelly1998ProportionalFairness}—that extend classical aggregation principles to cardinal utility spaces. These models formalize trade-offs among competing interests through axioms of efficiency, symmetry, and monotonicity, providing the mathematical scaffolding for balancing utilities among multiple actors.

More recently, research in \emph{computational social choice}~\citep{brandt2016handbook} has expanded these foundations by integrating algorithmic and complexity-theoretic perspectives across voting, fair allocation, and coalition formation. This body of work formalized how normative principles interact with computational constraints, revealing that properties such as fairness, consistency, and strategy-proofness often entail intrinsic computational trade-offs. Despite their theoretical depth, these frameworks typically assume fixed preferences and static aggregation rules, offering limited mechanisms for learning or adapting stakeholder preferences in dynamic, data-driven contexts. \\
\emph{Our framework operationalizes axiomatic bargaining and welfare operators over learned stakeholder utilities, unifying data-driven estimation with normative aggregation and introducing a robustness certificate schema to quantify the sensitivity of to perturbations.}
\paragraph{Multi-objective multi-agent systems} 
Research on multi-agent multi-objective decision-making examines how multiple agents optimize potentially conflicting objectives within shared environments. As surveyed by~\citet{ruadulescu2020multi}, this field unifies insights from multi-objective optimization, game theory, and multi-agent reinforcement learning (MARL). Existing methods, ranging from scalarization and Pareto-based search to policy ensembles and evolutionary optimization, enable trade-off exploration across competing objectives, yet often rely on fixed utility specifications and static criteria.
The authors identify several open challenges, including: (i) the lack of integration between optimization formulations across game-theoretic and learning paradigms; (ii) limited mechanisms for learning or eliciting agent utilities under partial observability or strategic manipulation; and (iii) the absence of standardized pipelines that couple preference modeling, negotiation, and execution. 
Parallel work in multi-criteria decision-making (MCDM) offers complementary aggregation frameworks grounded in multi-attribute utility theory. While effective for structuring trade-offs, these approaches typically rely on expert-driven weighting and static aggregation functions~\citep{taherdoost2023multi, chakraborty2023comprehensive}. Hybrid methods that integrate MCDM with learning or simulation~\citep{chakraborty2023comprehensive} partially address this rigidity but still lack principled mechanisms to align learned representations with normative fairness or robustness guarantees. Similarly, systematic reviews of sustainability-oriented MCDM frameworks ~\citep{boix2022systematic} highlight persistent limitations most notably (iv) limited multi-stakeholder integration, (v) overreliance on qualitative assessments, and (vi) the absence of mechanisms to penalize adverse or inequitable impacts. 

\emph{Our framework bridges game-theoretic and learning-based optimization paradigms within a standardized pipeline that robustly learns agent preferences, integrates multi-stakeholder perspectives, and can penalize inequitable impacts.}

\paragraph{Participatory AI as a paradigm.} \emph{Participatory AI} has emerged as a paradigm for integrating diverse stakeholder perspectives throughout the AI lifecycle, fostering context-dependent fairness and accountability \citep{birhane2022power}. This approach emphasizes collaboration across technical and non-technical domains~\citep{hossain2021towards, berditchevskaia2021participatory}, with applications spanning healthcare~\citep{donia2021co}, judicial systems~\citep{barabas2020studying}, civic engagement~\citep{arana2021citizen}, philanthropy~\citep{lee2019webuildai}, and urban planning~\citep{quan2019artificial}. Technical contributions include collective debiasing~\citep{chan2024group}, collaborative debugging~\citep{nakao2022toward}, preference-based ranking~\citep{cachel2024prefair}, and democratized ML workflows~\citep{zhang2023deliberating}. These efforts collectively represent a "participatory turn" in AI design~\citep{delgado2023participatory}. Despite this progress, significant barriers remain. Technical complexity, structural inequities, and power asymmetries risk reducing participation to "participation washing"~\citep{sloane2022participation}. \citet{maas2024beyond} argue for moving beyond participatory \emph{design} toward participatory \emph{systems} that embed stakeholder power in ongoing monitoring and adaptation. Similarly, \citet{feffer2023preference} note that current approaches often limit participation to static preference elicitation without operational integration. While participatory AI requires multi-dimensional solutions, technical limitations create a bottleneck to scalability~\citep{delgado2023participatory}, creating a vicious cycle in
which limited application hampers the development of richer participatory
mechanisms, which in turn further constrains their adoption
and broader impact.

\emph{Our framework operationalizes participatory AI as an end-to-end, scalable mechanism: stakeholder-informed reward signals are learned to generalize elicited preferences to new data, and decisions are selected transparently, embedding participation into the decision process itself rather than confining it to design-time elicitation.}\\

\section{The Participatory Learning and Decision Framework}

\label{sec:framework}

In traditional prediction-oriented decision-making systems, the goal is to recommend an action based on a set of features and a predicted outcome. 
In contrast, our framework reformulates this task as a \emph{multi-actor decision-making problem}, in which each actor holds distinct preferences over possible actions and outcomes, and context-aware recommendations must satisfy multiple evaluation criteria.

Before presenting the core components, it is important to emphasize that the framework is \emph{deliberately general and modular}. 
It provides a structured foundation for integrating stakeholder preferences into automated recommendation pipelines while remaining agnostic to specific modeling choices. 
Although we specify how actor-specific reward models are incorporated into the decision process, the concrete instantiation of individual components—such as function approximators, data augmentation strategies, reward elicitation methods, evaluation metrics, or decision functions—remains open by design to ensure adaptability across domains.

The effective use of the framework ultimately depends on the design of actor-specific reward signals, ideally grounded in authentic stakeholder input or objective impact measures. 
Addressing the broader challenge of aligning revealed and latent preferences lies beyond the present scope. 
Our focus is instead on \emph{simulating participatory decision-making at inference time}, given access to stakeholder-informed reward functions. 
The framework remains agnostic to the provenance of these signals while being compatible with future advances in preference elicitation and modeling.

\subsection{Problem Formulation}

\begin{definition}[Multi-Stakeholder Decision-Making Problem]\label{def:multi_actor_problem}
Let $\mathcal{I} = \{i_1,   i_2, \dots, i_n\}$ be a set of stakeholders or actors, where $|\mathcal{I}| = n$. 
We define a stakeholder as any real or symbolic entity that is influenced by the decisions suggested by the system and, therefore, holds a direct stake in its resulting outputs. 
The decision space consists of a set of possible actions $\mathcal{A} = \{a_1, a_2, \dots, a_k\}$, where $|\mathcal{A}| = k$, and a set of feasible outcomes $\mathcal{O} = \{o_1, o_2, \dots, o_m\}$, where $|\mathcal{O}| = m$.
Given a context $\boldsymbol{x} \in \mathcal{X}$ representing exogenous conditions (e.g., applicant attributes in a lending use-case), the system seeks to recommend an action $a \in \mathcal{A}$ that balances the preferences of all actors over the predicted outcomes.
\end{definition}

The framework flexibly accommodates both discrete and continuous action spaces $\mathcal{A}$ and outcome spaces $\mathcal{O}$, covering both classification- and regression-oriented formulations.\footnote{For continuous outcomes, direct integration or discretization can be employed as needed, yielding a unified treatment across problem types.}

\begin{assumption}[Discrete or Discretizable Action Space]\label{assumption:discrete_action}
While the framework is theoretically general and capable of handling continuous action spaces, the formal derivations and computational complexity analysis in this work assume that the action space $\mathcal{A}$ is either discrete or can be discretized into a finite representative set without significant loss of fidelity. This ensures that the system can efficiently evaluate and compare candidate actions across all components, maintaining computational tractability.\footnote{We remark that large or continuous action spaces may be addressed using established approximation techniques or optimization methods, depending on the specific structure of the problem and application domain.} This assumption aligns with real-world scenarios, where the set of feasible actions is typically finite or effectively constrained.
\end{assumption}

\subsection{Core Framework Components}

\subsubsection{Stakeholder preferences.}

Our framework incorporates actor-based signals that capture the desirability of each action-outcome pair for each stakeholder, leveraging feedback signals to guide decision-making aligned with heterogeneous priorities.
\begin{definition}[Reward-augmented dataset]
Consider a dataset of observed context–action–outcome triplets \(\mathcal{T} = \{(\boldsymbol{x}_t, a_t, o_t)\}_{t=1}^T\) and 
assume access to stakeholder-specific desirability annotations; then we can define the reward-augmented dataset as
\[
\mathcal{T}^+ = \{(\boldsymbol{x}_t, a_t, o_t, \mathbf{r}_t)\}_{t=1}^T,
\]
where \(\mathbf{r}_t = (r_{i,t})_{i \in \mathcal{I}} \in [0,1]^{|\mathcal{I}|}\) denotes the vector of reward signals provided by stakeholders \(i \in \mathcal{I}\) for the action–outcome pair \((a_t, o_t)\) under context \(\boldsymbol{x}_t\).
\end{definition}
\begin{definition}[Actor-Specific Reward Function]\label{def:actor_reward}
Given the multi-stakeholder decision-making problem described in Definition~\ref{def:multi_actor_problem}, 
each actor \( i \in \mathcal{I} \) is associated with a reward function
\[
R_i : \mathcal{X} \times \mathcal{A} \times \mathcal{O} \to [0,1],
\]
which assigns a normalized desirability score to each action–outcome pair \((a,o)\) under context \(\boldsymbol{x}\), reflecting the individual preferences of actor \( i \).\footnote{Normalization to \([0,1]\) is assumed without loss of generality to ensure comparability and bounded aggregation across actors.}
\end{definition}

\begin{assumption}[Surrogate Preference Model]\label{ass:surrogate_pref}
Each observed reward signal \( r_i(\boldsymbol{x},a,o) \) is treated as a sample from an elicitable surrogate function
\(
\hat{R}_i : \mathcal{X} \times \mathcal{A} \times \mathcal{O} \to [0,1],
\)
that approximates the actor’s latent reward function \( R_i \):
\[
r_i(\boldsymbol{x},a,o) \sim \hat{R}_i(\boldsymbol{x},a,o), 
\qquad 
\hat{R}_i(\boldsymbol{x},a,o) \approx R_i(\boldsymbol{x},a,o).
\]
The gap between \( \hat{R}_i \) and \( R_i \) depends on the expressiveness and reliability of the elicitation method 
(e.g., direct feedback, behavioral traces, or ranking-based heuristics). 
We treat this layer as exogenous to the framework and assume that reasonably accurate surrogates \(\hat{R}_i\) are available.
\end{assumption}
\begin{assumption}[Non-Adversarial Stakeholder Preferences]
\label{ass:nonadv}
Each actor \(i \in I\) reports reward signals \(r_i\) that truthfully reflect their preferences over action--outcome pairs, without strategic distortion.
\end{assumption}

Assumption \ref{ass:nonadv} complements Assumption \ref{ass:surrogate_pref} by clarifying that deviations between the true reward function \( R_i \) 
and its surrogate \( \hat{R}_i \) stem from \emph{epistemic uncertainty} (e.g., elicitation noise) rather than \emph{strategic manipulation}. 
Consequently, the stochasticity in \( r_i \sim \hat{R}_i \) captures informational, rather than behavioral, uncertainty. 
These assumptions hold in the general formulation but are \emph{relaxed analytically} in Section \ref{ref:robustness}
to accommodate adversarial and noisy preference settings. 

\begin{definition}[Stakeholder Reward Prediction Model]\label{def:reward_model}
For each stakeholder \( i \in \mathcal{I} \), a supervised model 
\[
q_i : \mathcal{X} \times \mathcal{A} \times \mathcal{O} \to [0,1]
\]
is trained on a representative subset \( \mathcal{T}^+_{\mathrm{sampled}} \subseteq \mathcal{T}^+ \) 
to approximate the elicited surrogate \( \hat{R}_i \) and enable context-aware estimation of desirability scores for unobserved action–outcome pairs.  
Under standard statistical learning assumptions~\citep{mohri2018foundations}, the learned predictors satisfy
\[
\mathbb{E}_{(\boldsymbol{x},a,o)\sim\mathcal{D}}
\big[\,|\, q_i(\boldsymbol{x},a,o) - \hat{R}_i(\boldsymbol{x},a,o)\,|\,\big] \le \epsilon_i,
\]
where \( \epsilon_i \ge 0 \) denotes the expected approximation error.
\end{definition}

\begin{remark}[Partial Reward Coverage and Generalization]\label{remark:partial_coverage}
In practice, the augmented dataset \( \mathcal{T}^+_{\mathrm{sampled}}  \) seldom covers all combinations of actions and outcomes.  
For small discrete spaces, exhaustive or near-exhaustive coverage allows direct construction of full matrices \( \mathbf{\hat{R}}_i(\boldsymbol{x}) \).  
For larger or continuous spaces, however, \( \mathcal{T}^+ \) typically provides a sampled subset, and the learned models \( q_i \) generalize to unobserved pairs under standard generalization guarantees~\citep{vapnik1999overview, hornik1989multilayer}.
\end{remark}

\begin{definition}[Predicted Reward Matrix]\label{def:predicted_reward_matrix}
Given a context \(\boldsymbol{x}\) and a learned model \( q_i \), 
the predicted reward matrix for actor \( i \) is defined as
\[
\mathbf{Q}_i(\boldsymbol{x}) \in [0,1]^{|\mathcal{A}|\times|\mathcal{O}|},
\qquad
[\mathbf{Q}_i(\boldsymbol{x})]_{a,o} := q_i(\boldsymbol{x},a,o),
\]
which enables counterfactual reasoning over the joint action–outcome space.
\end{definition}

\paragraph{Outcome Prediction Model.} 
As in standard supervised learning, we train an outcome predictor $f$ on the base dataset $\mathcal{T}$ to estimate outcomes from context--action pairs. The model supports both discrete and continuous settings:
\[
f : \mathcal{X} \times \mathcal{A} \to \Delta(\mathcal{O}) \quad \text{or} \quad f : \mathcal{X} \times \mathcal{A} \to \mathbb{R},
\]
where $\Delta(\mathcal{O})$ denotes the probability simplex over discrete outcomes.  Unlike traditional decision pipelines, $f$ does not directly inform the choice of action; instead, it serves as the foundation for computing expected actor-specific rewards.

\begin{definition}[Expected Reward Matrix] 
\label{def:Exp_r_tens}
Given a context $\boldsymbol{x} \in \mathcal{X}$, an outcome predictor $f$, and stakeholder-specific reward matrices $\mathbf{Q}_i(\boldsymbol{x}) \in [0,1]^{|\mathcal{A}| \times |\mathcal{O}|}$ for each $i \in \mathcal{I}$, we define the expected reward matrix:
\[
\mathbf{E}(\boldsymbol{x}) \in [0,1]^{|\mathcal{I}| \times |\mathcal{A}|},
\]
where each entry $\mathbf{E}_{i,a}(\boldsymbol{x})$ is the expected reward for stakeholder $i$ and action $a$, obtained by integrating over predicted outcomes:
\[
\mathbf{E}_{i,a}(\boldsymbol{x}) :=
\begin{cases}
\sum\limits_{o \in \mathcal{O}} \mathbf{Q}_i(\boldsymbol{x})_{a,o} \cdot f(o \mid \boldsymbol{x}, a), & \text{(discrete $\mathcal{O}$)}, \\
\int_{\mathcal{O}} \mathbf{Q}_i(\boldsymbol{x})_{a,o} \cdot f(o \mid \boldsymbol{x}, a)\, \mathrm{d}o, & \text{(continuous $\mathcal{O}$)}.
\end{cases}
\]

This matrix summarizes the expected utilities over actions and stakeholders for the given context and provides the input to the decision functions.
\end{definition}

\subsubsection{Decision strategies.}
Once the vectors of actor-specific expected rewards are computed
based on the predicted outcomes, a set of decision strategies can be
applied to derive the action suggested by the system. 
\begin{definition}[Decision Strategy]
A decision strategy is a mapping
\(
D : \mathcal{Z} \times \mathcal{P} \to \mathcal{Y},
\)
where \(\mathcal{Z}\) is the space of decision-relevant inputs, \(\mathcal{P}\) is a parameter space, and \(\mathcal{Y} \subseteq \Delta(\mathcal{A})\) is the set of valid decisions. Deterministic strategies correspond to the special case \(\mathcal{Y} = \mathcal{A} \subset \Delta(\mathcal{A})\).
\end{definition} 
In our setting, we instantiate \(\mathcal{Z} = [0,1]^{|\mathcal{I}| \times |\mathcal{A}|}\), representing the actor-specific expected reward matrix \(\mathbf{E}(\boldsymbol{x})\) for a given context \(\boldsymbol{x}\). We consider three broad types of strategies, namely (i) \emph{agent-agnostic strategies}, which ignore actor-specific reward signals and rely on outcome-based predictions, for example selecting the action with the highest predicted probability of a desirable outcome; (ii)
\emph{single-agent strategies}, which optimize for the interest of a specific actor \(i \in \mathcal{I}\) (e.g. selecting the action that maximizes their utility); (iii) \emph{multi-agent compromise strategies}, which aggregate heterogeneous preferences into a collective decision. These functions are designed to balance stakeholder utilities based on normative principles drawn from game theory, welfare economics, and multi-objective optimization.

\begin{definition}[Compromise Function]\label{def:compromise_function}
A \emph{compromise function} \( C_j \in \mathcal{C} \subseteq \mathcal{D} \) 
is a decision strategy that maps the expected reward matrix and a set of auxiliary parameters 
to a (possibly stochastic) distribution over actions:
\[
C_j : [0,1]^{|\mathcal{I}| \times |\mathcal{A}|} \times \mathcal{P} \to \Delta(\mathcal{A}).
\]
Each \( C_j \) is defined through a scoring function \(\Phi_j\) 
and a selection operator \(\Psi_j\) as follows:
\begin{equation*}
C_j(\mathbf{E}, \boldsymbol{p}) 
= \Psi_j\!\left( \left\{ \Phi_j(\mathbf{E}_{:,a}, \boldsymbol{p}) \right\}_{a \in \mathcal{A}} \right),
\end{equation*}
where \(\Phi_j : [0,1]^{|\mathcal{I}|} \times \mathcal{P} \to \mathbb{R}\) 
computes a scalar score for each action based on its vector of stakeholder utilities, 
and \(\Psi_j\) denotes a decision operator such as \(\arg\max\) or \(\mathrm{softmax}_\tau\).
\end{definition}

\subsubsection{Decision Evaluation.}
To compare decision strategies across context-specific multiple evaluation criteria, we define a normalized composite score that aggregates raw metric-specific performances into a unified objective.
\begin{definition}[Composite Evaluation Score]\label{def:score_base}
Consider a set of evaluation metrics \(\mathcal{M} = \{M_1, \dots, M_z\}\), each inducing a raw score:
\[
\mathcal{S}_h^{\mathrm{raw}}(\mathbf{E}; D) 
:= \frac{1}{T^{\gamma}} 
\sum_{t=1}^{T} \sum_{a \in \mathcal{A}} \sum_{o \in \mathcal{O}} 
\Theta\!\big(D(\boldsymbol{x}_t), M_h \mid \boldsymbol{x}_t, a, o\big),
\]
where \(\Theta(\cdot)\) captures the metric-specific contribution for triplet \((\boldsymbol{x}_t, a, o)\), 
and \(\gamma \in \{0,1\}\) distinguishes between averaging metrics (\(\gamma=1\)) and summing metrics (\(\gamma=0\)).

To ensure comparability across metrics, each raw score is normalized over the set of candidate strategies \(\mathcal{D}\):
\[
\tilde{\mathcal{S}}_h(\mathbf{E}; D) 
:= \frac{
\mathcal{S}_h^{\mathrm{raw}}(\mathbf{E}; D) 
- \min\limits_{D' \in \mathcal{D}} \mathcal{S}_h^{\mathrm{raw}}(\mathbf{E}; D')
}{
\max\limits_{D' \in \mathcal{D}} \mathcal{S}_h^{\mathrm{raw}}(\mathbf{E}; D')
- \min\limits_{D' \in \mathcal{D}} \mathcal{S}_h^{\mathrm{raw}}(\mathbf{E}; D')
}.
\]
This standardization maps all metric values to the interval \([0,1]\), 
where \(1\) denotes optimal and \(0\) worst performance, according to the metric’s semantic orientation.
\end{definition}

Given a weight vector \(\boldsymbol{w} = (w_1, \dots, w_z) \in \Delta(\mathcal{M})\), the composite score is:
\begin{equation*}
\label{eq:score_general}
\mathcal{S}(\mathbf{E}; D) := \sum_{h=1}^{z} w_h \cdot \tilde{\mathcal{S}}_h(\mathbf{E}; D),
\end{equation*}

from which we can compute 
\(
D^* := \arg\max_{D \in \mathcal{D}}\mathcal{S}(\mathbf{E}; D) 
\)

\begin{remark}
In practice, the composite score \( \mathcal{S}(D) \) is estimated via a standard validation protocol (such as \(k\)-fold cross-validation) while jointly tuning the outcome model \( f \) and the stakeholder-specific reward models \( q_i \). 
This procedure enhances robustness to overfitting and enables out-of-sample selection of the optimal decision strategy \( D^* \) based on empirical generalization performance.

\end{remark}
\subsubsection{Best action selection.}
At inference, given a new context \( \boldsymbol{x}' \in \mathcal{X} \), the system computes the expected reward matrix \( \mathbf{E}(\boldsymbol{x}') \) by combining the outcome predictor \( f \) and stakeholder models \( q_i \), and selects the final decision via the optimal strategy:
\(
a^* = D^*(\mathbf{E}(\boldsymbol{x}'); \boldsymbol{p}),
\)
where \( D^* \in \mathcal{D} \) is the previously selected optimal decision rule and \( \boldsymbol{p} \in \mathcal{P} \) are its associated parameters.\\

\subsection{Computational Complexity}

\begin{theorem}[Additional Computational Overhead]
Under Assumption \ref{assumption:discrete_action}, relative to a baseline outcome-prediction system, the additional computational overhead introduced by the frameworks, added on top of the base cost, is:

\textbf{Offline (per cross-validation run):}
\[
O\Big( |\mathcal{I}| \cdot \big( |G_q| \cdot c_{\mathrm{train}}^q + T_{\mathrm{val}} \cdot |\mathcal{A}| \cdot (c_{\mathrm{inf}}^q + |\mathcal{D}|) \big) + T_{\mathrm{val}} \cdot |\mathcal{D}| \cdot |\mathcal{M}| \Big)
\]

\textbf{Online (per test instance):}
\[
\begin{cases}
O\big( |\mathcal{A}| \cdot |\mathcal{I}| \cdot (c_{\mathrm{inf}}^q + 1) \big) & \text{if using preselected } D^* \\
O\big( |\mathcal{A}| \cdot |\mathcal{I}| \cdot (c_{\mathrm{inf}}^q + |\mathcal{D}|) \big) & \text{if evaluating all } D \in \mathcal{D}
\end{cases}
\]

\noindent where \( |G_q| \) is the size of the hyperparameter grid for stakeholder models \( q_i \), \( T_{\mathrm{val}} \) is the validation set size, \( c_{\mathrm{train}}^q, c_{\mathrm{inf}}^q \) are training and inference cost per \( q_i \).
\end{theorem}

\begin{remark}[Real-world settings] In real-world deployments, where the number of actors, actions, strategies, and metrics is typically moderate, the added overhead remains practically tractable.
\end{remark}
\begin{remark}[Effect of Non-Convexity]
Even if some compromise objectives $\Phi_j$ are non-convex and solved heuristically, under Assumption~1 the worst-case per-context cost remains $\mathcal{O}(|\mathcal{A}|)$, leaving the overall asymptotic complexity unchanged.
\end{remark}

\section{Experiments}
\label{sec:experiments}
\subsection{Scope of Empirical Validation}
Although the core contribution of this paper is the formal specification of the framework, we complement it with illustrative experiments to explore its behavior and applicability across real-world decision-making settings. In particular, we provide empirical evidence of the framework’s potential in two representative high-stakes domains, demonstrating its flexibility, utility, and capacity to support stakeholder-aligned decisions. Additionally, we conduct ablation analyses to examine how variations in reward structure, training sample size, and predictive model capacity influence system performance across a range of evaluation metrics.
Notably, our experiments are designed to showcase how the framework enables modular evaluation and accountable decision selection under heterogeneous stakeholder utilities. As it is agnostic to the underlying predictor, we do not compare against methods that modify the learning pipeline. The framework already subsumes common aggregation strategies: the weighted-sum scalarization appears as the utilitarian compromise rule $\Phi_{\text{sum}} \in \mathcal{D}$, and post-hoc fairness corrections correspond to constraint-enforced variants of $\Phi_j$. Hence, our analysis inherently contrasts these baseline behaviors alongside single-actor strategies and predictive-only baselines
with richer compromise operators, namely Maximin\citep{Wald1945Minimax}, Kalai--Smorodinsky \citep{kalai1975other}, Nash Bargaing Solution \citep{nash1950bargaining}, Proportional Fairness \citep{Kelly1998ProportionalFairness} and Compromise Programming \citep{Zeleny1973Compromise}, within a unified evaluation pipeline.
  
\begin{figure*}[t]
    \centering

    \begin{subfigure}[t]{0.9\linewidth}
        \centering
        \includegraphics[width=\linewidth]{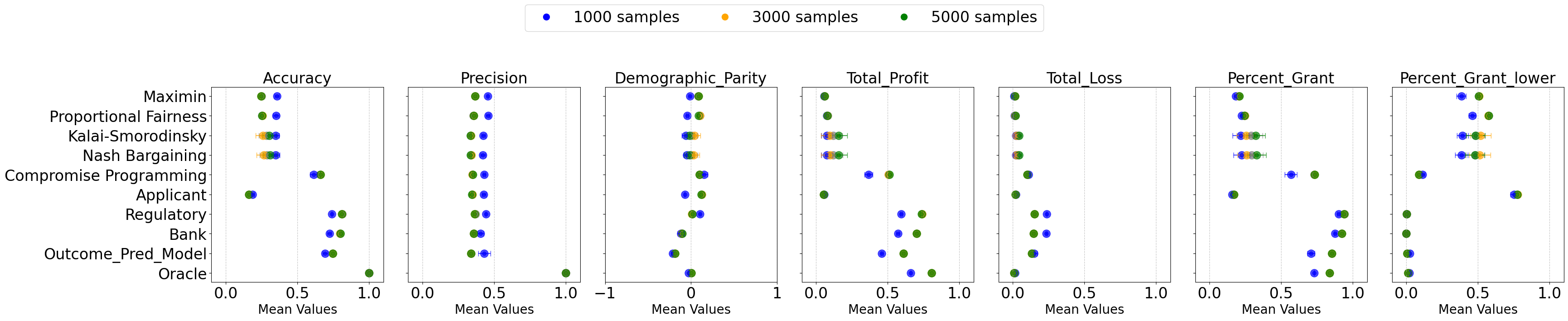}
        \caption{Lending scenario; varying reward structure.}
        \label{fig:abl_reward}
    \end{subfigure}
    \vspace{1em}

    \begin{subfigure}[t]{0.9\linewidth}
        \centering
        \includegraphics[width=\linewidth]{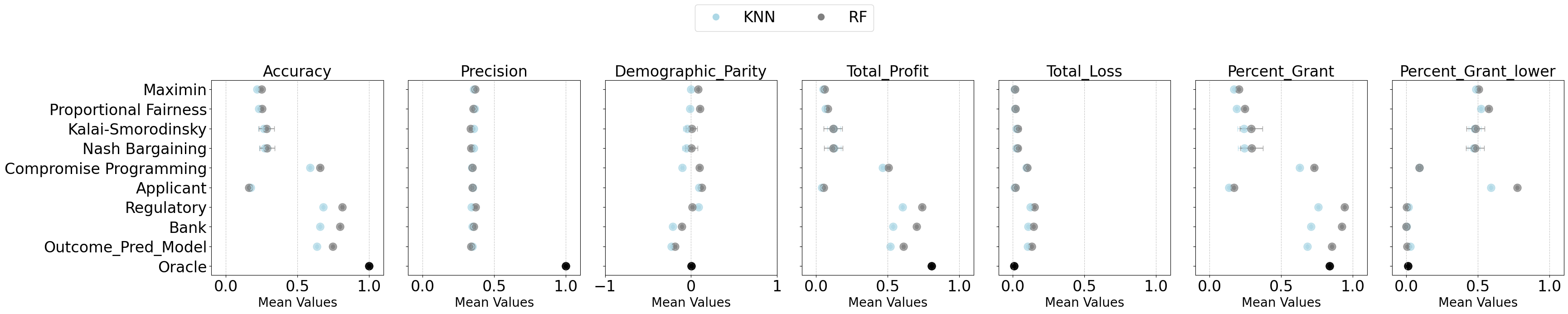}
        \caption{Lending scenario; varying predictive model.}
        \label{fig:abl_predmodel}
    \end{subfigure}
    \vspace{1em}

    \begin{subfigure}[t]{0.9\linewidth}
        \centering
        \includegraphics[width=\linewidth]{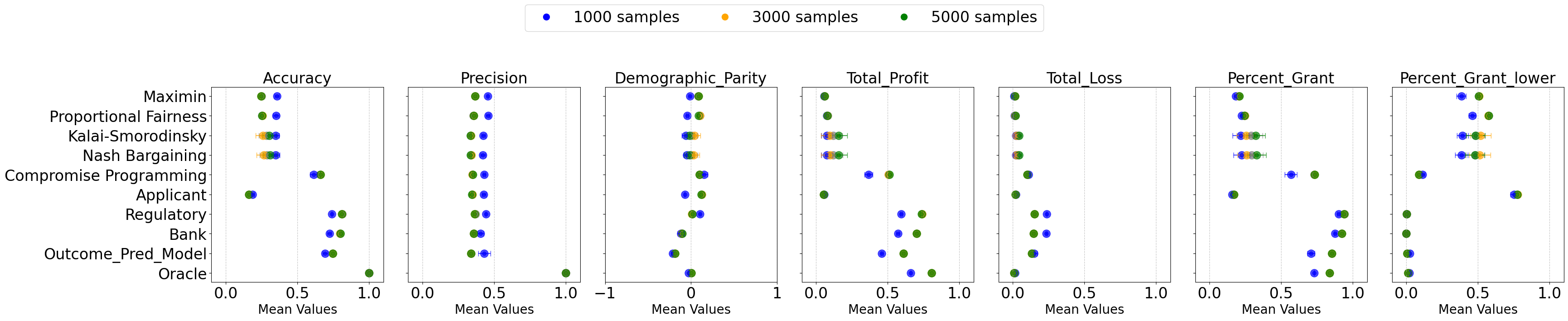}
        \caption{Lending scenario; varying training sample size.}
        \label{fig:abl_sample}
    \end{subfigure}
    \vspace{1em}

    \begin{subfigure}[t]{0.9\linewidth}
        \centering
        \includegraphics[width=\linewidth]{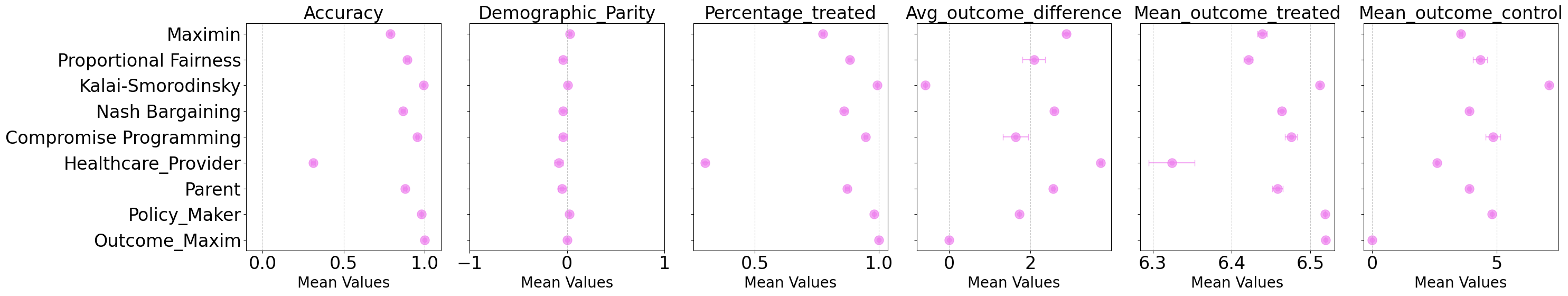}
        \caption{Healthcare scenario.}
        \label{fig:abl_health}
    \end{subfigure}

    \caption{
    Comparison of decision strategy performance across evaluation metrics under different experimental conditions. 
    Subfigures~(a)--(c) report results for the lending scenario under variations in reward structure, predictive model, and sample size, respectively; subfigure~(d) presents analogous results for the healthcare scenario. 
    Each panel displays mean and standard deviation of test values for multiple compromise operators and baseline strategies, illustrating trade-offs between predictive accuracy, fairness, and welfare efficiency across configurations.
    }
    \Description{
    Vertically arranged subfigures comparing decision strategy performance across four experimental conditions. 
    The first three subfigures correspond to the lending scenario (varying reward structure, predictive model, and training sample size), and the last to the healthcare scenario. 
    Each subplot shows evaluation metrics such as accuracy, precision, demographic parity, profit, loss, or treatment outcomes, reported as mean and standard deviation across decision operators. 
    The figure highlights how different compromise formulations balance fairness, performance, and welfare efficiency.
    }
    \label{fig:abl_all_vertical}
\end{figure*}

\subsection{Real-World Use Cases}
We present two use cases, namely a loan approval scenario (multi-classification) and a health treatment selection scenario (causal inference regression). In both experimental settings, reward structures are predefined using prototypical stakeholder heuristics with added uniform noise to simulate real-world variability and test model robustness. 
\subsubsection*{Use Case 1: Lending Scenario}

We apply the framework to real-world data from the publicly available Lending Club dataset, structured as a 
$3 \times 3$ problem where the decision space $\mathcal{A}$ includes three actions ("Grant," "Grant lower amount," "Not Grant") and the outcome space $\mathcal{O}$ consists of three repayment states ("Fully Repaid," "Partially Repaid," "Not Repaid"). Context features include applicant-specific attributes such as credit score, income, financial history, and demographics. This setting models three key stakeholder groups, namely 
(i) \textit{Bank}, seeking to maximize profitability, with rewards tied to repayment outcomes; (ii)  \textit{Applicant}, prioritizing loan access, with rewards reflecting the utility of approval balanced against repayment obligations; and (ii) \textit{Regulatory Body}, aiming to ensure financial stability and inclusivity, placing value on equitable credit access for vulnerable groups.
Beyond the baseline strategies, we benchmark each compromise function against an \textit{Oracle} strategy that simulates an idealized bijective mapping between outcomes and optimal actions. 
Performance is evaluated using standard fairness (\emph{Demographic\_Parity}) and accuracy metrics, alongside domain-specific measures such as profit (\emph{Total\_Profit}) and loss (\emph{Total\_Loss}) percentages, computed relative to the Oracle benchmark. 
We also report the proportion of fully or partially granted loans (\emph{Percent\_Grant} and \emph{Percent\_Grant\_Lower}), corresponding to the actions recommended by each strategy. 
Both the outcome prediction model and the actor-specific reward models are implemented as Random Forests, selected for their robustness to non-differentiable reward signals and wide practical adoption. 
For ablation analysis, we additionally include a simpler $k$-Nearest Neighbor baseline.

\subsubsection*{Use Case 2: Healthcare Scenario}

For the healthcare scenario, we use the first realization of the Infant Health and Development Program (IHDP) dataset, as introduced by \citet{hill2011bayesian} and widely employed in causal inference research \citep{shalit2017estimating, louizos2017causal, yao2018representation}. This dataset originates from a randomized experiment studying the effect of home visits on infant cognitive test scores. The decision space comprises two actions (treatment vs. no treatment), with outcomes given by continuous cognitive scores. The context vector includes 25 binary and continuous features describing child and family characteristics.

This scenario illustrates the framework’s applicability to causal inference tasks, modeling three stakeholder groups, namely
(i) \textit{Healthcare Provider}, aiming to improve patient outcomes while managing costs;
(ii) \textit{Policy Maker}, seeking to maximize societal benefit and ensure equity across demographic groups; and 
(iii) \textit{Parent}, valuing direct improvements in child's well-being.

In addition to individual maximization baselines, we include a strategy (\emph{Outcome\_Maxim}) that maximizes overall cognitive scores which naturally recommends treatment for all potential patients. Case-specific evaluation metrics include the mean outcome value for the treated (\emph{Mean\_outcome\_treated}) versus the control group (\emph{Mean\_outcome\_control})  and their average absolute difference \\(\emph{Avg\_outcome\_difference}). To estimate the Conditional Average Treatment Effect (CATE), we use an X-regressor meta-learner \citep{chen2020causalml} with XGBRegressor as the base learner.

\subsection{Discussion of Key Insights}
Beyond identifying the optimal strategy, the framework enables systematic comparison of decision functions across multiple evaluation metrics. To this aim, Figure 1 shows raw framework outputs that a hypothetical user interface could display to transparently reveal trade-offs across metrics and decision functions. 
Figures~\ref{fig:abl_reward} and \ref{fig:abl_predmodel} provide a comprehensive empirical breakdown of how variations in reward structures and outcome model complexity shape decision function performance across predictive, fairness, and case-specific metrics in the lending scenario. All results are averaged over four random seeds and benchmarked against an \textit{Oracle} strategy to reveal upper-bound potential for outcome prediction.
A key observation across all experiments is that baseline strategies tend to optimize classical predictive metrics, notably accuracy, but underperform in fairness dimensions such as \textit{Demographic\_Parity} and in the diversity of actions recommended. More broadly, the observed performance differences between compromise-based strategies and single-actor or single-objective baselines underscore the inherent limitations of narrowly optimized systems: while such baselines may perform well on isolated metrics, they consistently underperform when assessed across a broader range of evaluation dimensions, particularly those capturing equity and diversity. Critically, the magnitude of the performance gap between narrow-objective solutions and multi-actor compromise approaches reveals the limited capacity of traditional single-perspective systems to reconcile competing priorities — and, by extension, their diminished ability to deliver socially informed, context-sensitive, and accountable decisions. In particular, Figure~\ref{fig:abl_reward} reveals how shifting the underlying actor reward functions — from balanced to more self-interested formulations\footnote{In \emph{Strictest Strategy} setting in Figure \ref{fig:abl_reward}, the \emph{Bank} values loan approvals only when full repayment is expected, while the \emph{Applicant} prioritizes loan approval regardless of repayment likelihood.} — sharpens the performance–fairness trade-off. This underscores the subtle but powerful role that reward signals plays in shaping aggregate system behavior.

Figure~\ref{fig:abl_predmodel} and ~\ref{fig:abl_sample}  together allow observing the impact of increasing predictive power — whether through greater model expressiveness or larger training sample size — on decision strategy performance. The figures show that while improvements in performance-oriented metrics for single-objective strategies are evident (see \emph{Accuracy} and \emph{Total\_Profit}), they are not accompanied by proportionally large gains in other dimensions, nor do they substantially and unambiguously alter the relative performance of compromise-based strategies. This emphasizes, on one hand, the need to evaluate systems across a broader set of dimensions beyond predictive accuracy alone, and on the other hand, that when balancing diverse stakeholder preferences, simply optimizing the predictive power of the underlying model remains insufficient.
Finally, Figure \ref{fig:abl_health} presents the results for the healthcare scenario, illustrating how the proposed framework can be used to analyze the distribution of treatment effects across patients based on treatment assignment decisions and expected treatment outcomes. This experiment highlights the framework’s capacity to jointly evaluate causal, fairness, and welfare dimensions of decision-making. Beyond standard causal metrics—such as the average treatment effect and group-specific outcome differences—the analysis also reveals how alternative decision strategies distribute treatment benefits and burdens across population subgroups.

Taken together, these insights highlight the versatility of the proposed framework across domains and its potential to enable more stakeholder-sensitive and context-aware accountable decisions, even when operating on identical predictive inputs and without relying on pre-imposed or rigid fairness constraints. 

\section{Robustness Certificates}
\label{ref:robustness}
In this section, we aim to extend the framework to settings where reported rewards may be perturbed by noise or strategic manipulation. 
We model these deviations as norm-bounded perturbations of the expected reward tensor~$\mathbf{E}$ and derive analytical robustness certificates that quantify the worst-case degradation of collective decisions under such disturbances. To decouple robustness analysis from the particular source of distortions in stakeholder rewards, we model perturbations directly in $\mathbf{E}$. This abstraction supports worst-case robustness analysis under both adversarial and approximation-driven deviations.
Throughout this section, $\mathbf{E}\in[0,1]^{|\mathcal{X}|\times|\mathcal{I}|\times|\mathcal{A}|}$ denotes the expected-reward tensor with entries $\mathbf{E}_{x,i,a}$. We write $B_\infty(\mathbf{E},\delta):=\{\mathbf{E}':\|\mathbf{E}'-\mathbf{E}\|_\infty\le\delta\}$. We use $\nabla_V\mathcal{S}_j^\tau$ to indicate the restriction of the gradient to the coordinates with $i\in V$.

\begin{definition}[Coalition-Based Perturbation Set]
\label{def:pert_set}
For coalition $V\subseteq \mathcal{I}$ where $v = |V|$, define the $\ell_\infty$-constrained perturbation set:
\begin{equation*}
\mathcal{U}_\infty(\delta) := \left\{
\Delta \;\middle|\;
\begin{array}{l}
\|\Delta\|_\infty \le \delta, \\
\Delta_{x,i,a} = 0 \quad \forall i \notin V, \\
\mathbf{E} + \Delta \in [\mu, 1-\mu]^{|\mathcal{X}| \times |\mathcal{I}| \times |\mathcal{A}|}
\end{array}
\right\}.
\end{equation*}
The entrywise cap $\|\Delta\|_\infty \leq \delta$ models uncoordinated per-actor manipulation conservatively. To ensure bounded curvature, we require:
\begin{equation*}
\delta < \min_{\substack{x \in \mathcal{X} \\ i \in V \\ a \in \mathcal{A}}} \min\left( \mathbf{E}_{x,i,a} - \mu,\; 1 - \mathbf{E}_{x,i,a} - \mu \right)
\end{equation*}
with $\mu \in (0, \tfrac{1}{2})$ enforcing an interior tube $[\mu, 1-\mu]$\footnote{We clip entries of the expected reward tensor into $[\mu,1-\mu]$ to avoid boundary effects (e.g., for log-based rules) and control curvature.}.

\end{definition}

\begin{assumption}[Softmax--Argmax Bias]\label{ass:bias}
There exists $\kappa_{C_j}(v)>0$ such that for all $\mathbf{E}' \in B_\infty(\mathbf{E},\delta)$,
\(
\big|\mathcal{S}_j^{\tau}(\mathbf{E}')-\mathcal{S}_j^{0}(\mathbf{E}')\big|
\le \kappa_{C_j}(v)\,\tau.
\)
\end{assumption}

\begin{theorem}[Composite Score Robustness Certificate schema]
\label{thm:robust_bound}
Let $v = |V|$ and assume:
\begin{enumerate}
    \item $\mathbf{E} \in [\mu,1-\mu]^{|\mathcal{X}| \times |\mathcal{I}| \times |\mathcal{A}|}$
    \item $\Delta \in \mathcal{U}_\infty(\delta)$ (Def.~\ref{def:pert_set})
    \item $C_j^\tau = (\Phi_j, \mathrm{softmax}_\tau)$ with $\Phi_j \in C^1$, and $C_j^0 = (\Phi_j, \arg\max)$
    \item  $\mathcal{S}_j^\tau$ is $\big(\beta_{C_j}(v)/(\tau^2\mu^2)\big)$-smooth under $\ell_\infty$.

\end{enumerate}
For brevity, we denote 
\(\mathcal{S}_j^{\tau}(\mathbf{E}) := \mathcal{S}_j^{\tau}(\mathbf{E}; C_j^{\tau})\) and 
\(\mathcal{S}_j^{0}(\mathbf{E}) := \mathcal{S}_j^{0}(\mathbf{E}; C_j^{0})\),
with the same convention for \(\mathbf{E}' \in B_\infty(\mathbf{E}, \delta)\). \\

Then the worst-case sharp evaluation satisfies:
\[
\inf_{\Delta \in \mathcal{U}_\infty(\delta)} \mathcal{S}_j^{0}(\mathbf{E} + \Delta)
\;\ge\;
\mathsf{RLB}_v^\tau(\mathbf{E}; C_j)
\]
where:

\begin{align*}
&\mathsf{RLB}_v^\tau(\mathbf E;\,C_j)
=
  \mathcal S_j^{\tau}\!\bigl(\mathbf E\bigr)
  \;-\;
  \mathcal E_v^\tau(\mathbf E;\,C_j)
  \;-\;
  \mathcal B_v^\tau(C_j),
\\[6pt]
&\mathcal E_v^\tau(\mathbf E;\,C_j)
=
  \underbrace{\delta\,\bigl\|\nabla_V\mathcal S_j^{\tau}\bigr\|_1}_{\text{gradient term}}
  \;+\;
  \underbrace{\frac{\delta^2\,\beta_{C_j}(v)}{2\,\tau^{2}\mu^{2}}}_{\text{curvature term}},
\\[6pt]
&\mathcal B_v^\tau(C_j)
:=\;
  \sup_{\mathbf E' \in B_\infty(\mathbf E,\delta)}
  \bigl|\,
    \mathcal S_j^{\tau}(\mathbf E')
    -
    \mathcal S_j^{0}(\mathbf E')
  \bigr|
  \;\le\;
  \kappa_{C_j}(v)\,\tau.
\end{align*}

The bound decomposes into:
\begin{itemize}
    \item[(i)] a \emph{first-order gradient term}, whose magnitude increase at most linearly with the coalition size \(v\), under bounded per-actor sensitivity.
    \item[(ii)] a \emph{second-order curvature penalty}, that accounts for nonlinearity in the decision rule \(C_j\). It depends on both the coalition size \(v\) and the geometry of the aggregation function \(\Phi_j\), and is controlled by the curvature constant \(\beta_{C_j}(v)\);
    \item[(iii)] a \emph{bias term}, measuring the approximation gap induced by softmax relaxation,  where \(B_\infty(\mathbf{E}, \delta)\) denotes the \(\ell_\infty\)-ball of radius \(\delta\) around \(\mathbf{E}\), i.e. $\{\mathbf E' : \|\mathbf E'-\mathbf E\|_\infty\le\delta\}$.
\end{itemize}
\end{theorem}
\begin{remark}
Theorem 2 provides a general robustness certificate schema, parameterized by curvature and smoothness constants that capture the local geometry of the compromise operator.
We intentionally keep these constants abstract to maintain generality across operator families. 
\end{remark}
\begin{lemma}[Optimal Temperature]
\label{lemma:temp}
The optimal temperature \(\tau^*\) minimizing the surrogate approximation–robustness tradeoff in Theorem~\ref{thm:robust_bound} is:
\[
\tau^* = \left(\frac{\delta^2 \beta_{C_j}(v)}{\kappa_{C_j}(v) \mu^2}\right)^{1/3}
\]
This balances the second-order curvature term and the linear bias term.
\end{lemma}

\begin{corollary}[Uniform Bias Control at Optimal Temperature]
\label{cor:uniform_control}
Let $\tau^*$ be the temperature from Lemma~\ref{lemma:temp}. Then the smooth-to-sharp approximation error is uniformly bounded across $B_\infty(\mathbf{E}, \delta)$:
\[
\sup_{\mathbf{E}' \in B_\infty(\mathbf{E}, \delta)}
\left|
\mathcal{S}_j^{\tau^*}(\mathbf{E}') - \mathcal{S}_j^{0}(\mathbf{E}')
\right|
\le
\kappa_{C_j}(v)\, \tau^*
\]
\end{corollary}

\begin{lemma}[Global Ranking Consistency]
\label{lem:global_consistency}
Let $\mathcal R=\{C_1,\dots,C_m\}$ denote a set of $m$ aggregation rules and let
their sharp scores be given by 
$S_g^{0}:=\mathcal S^{0}_g(\mathbf E)$. 
Assume the sharp evaluation induces a strict ordering
$S^{0}_{1} > S^{0}_{2} > \dots > S^{0}_{m}$.
For each rule $g$ let $\tau_g^{*}$ be the optimal temperature from
Lemma~\ref{lemma:temp} and define 
\(
  \epsilon_g := \kappa_{C_g}(v)\,\tau_g^{*}\quad(\forall g\le m).
\) 
If
\[
\displaystyle 
\min_{g<h}\bigl|S^{0}_g - S^{0}_h\bigr|
\;>\;
\max_{g<h}(\epsilon_g+\epsilon_h)
\]

then for every perturbed tensor
$\mathbf E'\in B_\infty(\mathbf E,\delta)$ the smooth scores preserve
the entire ranking:
\[
  S^{\tau^{*}_1}_1(\mathbf E')
  > S^{\tau^{*}_2}_2(\mathbf E')
  > \dots
  > S^{\tau^{*}_m}_m(\mathbf E').
\]
\end{lemma}

\paragraph{Interpretation.} 
Ranking consistency requires that sharp score gaps exceed the combined approximation errors.
Each error term \(\epsilon_g = \kappa_{C_g}(v)\, \tau_g^*\) bounds the smooth-to-sharp discrepancy under perturbations, increasing with coalition size \(v\) and radius \(\delta\). The optimal temperature \(\tau_g^*\) minimizes this bound, enhancing the likelihood of preserving the sharp ranking under adversarial shifts.

\section{Conclusion and Future Work}
\label{sec:conclusion}
Traditional data-driven systems typically recommend actions based solely on predictive accuracy or fixed performance metrics, which, even when constrained by specific fairness criteria, often overlook the diverse and sometimes conflicting preferences of stakeholders. In this work, we introduced a 
 flexible and model-agnostic learning framework that integrates heterogeneous stakeholder interests into the automated decision-making process, effectively simulating a participatory process whenever new decisions must be made. By combining actor-specific reward modeling with compromise strategies, the framework enables the balancing of competing priorities while optimizing user-defined evaluation metrics across domains.
 
Our experiments show that the framework flexibly improves context-awareness, accountability, and transparency in decision-support, letting practitioners trace how stakeholder preferences shape decisions and justify trade-offs, without imposing rigid fairness constraints or altering base predictors. Future work includes: (i) real-world deployments with active preference elicitation and user studies; (ii) instantiating constants for specific compromise families and identifying conditions for tight robustness bounds; and (iii) extending the framework to handle evolving stakeholder preferences over time.


\bibliographystyle{ACM-Reference-Format} 
\bibliography{main}

@article{louizos2017causal,
  title={Causal effect inference with deep latent-variable models},
  author={Louizos, Christos and Shalit, Uri and Mooij, Joris M and Sontag, David and Zemel, Richard and Welling, Max},
  journal={Advances in neural information processing systems},
  volume={30},
  year={2017}
}

@article{hill2011bayesian,
  title={Bayesian nonparametric modeling for causal inference},
  author={Hill, Jennifer L},
  journal={Journal of Computational and Graphical Statistics},
  volume={20},
  number={1},
  pages={217--240},
  year={2011},
  publisher={Taylor \& Francis}
}

@inproceedings{shalit2017estimating,
  title={Estimating individual treatment effect: generalization bounds and algorithms},
  author={Shalit, Uri and Johansson, Fredrik D and Sontag, David},
  booktitle={International conference on machine learning},
  pages={3076--3085},
  year={2017},
  organization={PMLR}
}

@article{yao2018representation,
  title={Representation learning for treatment effect estimation from observational data},
  author={Yao, Liuyi and Li, Sheng and Li, Yaliang and Huai, Mengdi and Gao, Jing and Zhang, Aidong},
  journal={Advances in neural information processing systems},
  volume={31},
  year={2018}
}

@article{quan2019artificial,
  title={Artificial intelligence-aided design: Smart design for sustainable city development},
  author={Quan, S. J. and Park, J. and Economou, A. and Lee, S.},
  journal={Environment and Planning B: Urban Analytics and City Science},
  volume={46},
  number={8},
  pages={1581--1599},
  year={2019}
}

@inproceedings{sloane2022participation,
  title={Participation is not a design fix for machine learning},
  author={Sloane, M. and Moss, E. and Awomolo, O. and Forlano, L.},
  booktitle={Equity and Access in Algorithms, Mechanisms, and Optimization},
  pages={1--6},
  year={2022}
}

@inproceedings{selbst2019fairness,
  title={Fairness and abstraction in sociotechnical systems},
  author={Selbst, Andrew D and Boyd, Danah and Friedler, Sorelle A and Venkatasubramanian, Suresh and Vertesi, Janet},
  booktitle={Proceedings of the conference on fairness, accountability, and transparency},
  pages={59--68},
  year={2019}
}

@article{gerdon2022social,
  title={Social impacts of algorithmic decision-making: A research agenda for the social sciences},
  author={Gerdon, Frederic and Bach, Ruben L and Kern, Christoph and Kreuter, Frauke},
  journal={Big Data \& Society},
  volume={9},
  number={1},
  pages={20539517221089305},
  year={2022},
  publisher={SAGE Publications Sage UK: London, England}
}

@article{donia2021co,
  title={Co-design and ethical artificial intelligence for health: An agenda for critical research and practice},
  author={Donia, J. and Shaw, J. A.},
  journal={Big Data \& Society},
  volume={8},
  number={2},
  pages={20539517211065248},
  year={2021},
  publisher={SAGE Publications}
}

@article{arana2021citizen,
  title={Citizen participation and machine learning for a better democracy},
  author={Arana-Catania, M. and Lier, F. A. V. and Procter, R. and Tkachenko, N. and He, Y. and Zubiaga, A. and Liakata, M.},
  journal={Digital Government: Research and Practice},
  volume={2},
  number={3},
  pages={1--22},
  year={2021}
}

@inproceedings{barabas2020studying,
  title={Studying up: reorienting the study of algorithmic fairness around issues of power},
  author={Barabas, C. and Doyle, C. and Rubinovitz, J. B. and Dinakar, K.},
  booktitle={Proceedings of the 2020 Conference on Fairness, Accountability, and Transparency},
  pages={167--176},
  year={2020}
}

@book{berditchevskaia2021participatory,
  title={Participatory AI for humanitarian innovation},
  author={Berditchevskaia, A. and Malliaraki, E. and Peach, K.},
  year={2021},
  publisher={Nesta},
  address={London}
}

@article{gerdes2022participatory,
  title={A participatory data-centric approach to AI Ethics by Design},
  author={Gerdes, A.},
  journal={Applied Artificial Intelligence},
  volume={36},
  number={1},
  pages={2009222},
  year={2022}
}

@inproceedings{delgado2023participatory,
  title={The participatory turn in ai design: Theoretical foundations and the current state of practice},
  author={Delgado, Fernando and Yang, Stephen and Madaio, Michael and Yang, Qian},
  booktitle={Proceedings of the 3rd ACM Conference on Equity and Access in Algorithms, Mechanisms, and Optimization},
  pages={1--23},
  year={2023}
}

@article{schoeffer2022relationship,
  title={On the relationship between explanations, fairness perceptions, and decisions},
  author={Schoeffer, Jakob and De-Arteaga, Maria and Kuehl, Niklas},
  journal={arXiv preprint arXiv:2204.13156},
  year={2022}
}

@inproceedings{dodge2019explaining,
  title={Explaining models: an empirical study of how explanations impact fairness judgment},
  author={Dodge, Jonathan and Liao, Q Vera and Zhang, Yunfeng and Bellamy, Rachel KE and Dugan, Casey},
  booktitle={Proceedings of the 24th international conference on intelligent user interfaces},
  pages={275--285},
  year={2019}
}

@article{ramachandranpillai2023fairxai,
  title={FairXAI-A Taxonomy and Framework for Fairness and Explainability Synergy in Machine Learning},
  author={Ramachandranpillai, Resmi and Baeza-Yates, Ricardo and Heintz, Fredrik},
  journal={Authorea Preprints},
  year={2023},
  publisher={Authorea}
}

@article{jain2020biased,
  title={Biased models have biased explanations},
  author={Jain, Aditya and Ravula, Manish and Ghosh, Joydeep},
  journal={arXiv preprint arXiv:2012.10986},
  year={2020}
}

@inproceedings{goel2021importance,
  title={The importance of modeling data missingness in algorithmic fairness: A causal perspective},
  author={Goel, N. and Amayuelas, A. and Deshpande, A. and Sharma, A.},
  booktitle={Proceedings of the AAAI Conference on Artificial Intelligence},
  volume={35},
  number={9},
  pages={7564--7573},
  year={2021}
}

@article{nakao2022toward,
  title={Toward involving end-users in interactive human-in-the-loop AI fairness},
  author={Nakao, Yuri and Stumpf, Simone and Ahmed, Subeida and Naseer, Aisha and Strappelli, Lorenzo},
  journal={ACM Transactions on Interactive Intelligent Systems},
  volume={12},
  number={3},
  pages={1--30},
  year={2022},
  publisher={ACM New York, NY}
}

@article{hossain2021towards,
  title={Towards a New Participatory Approach for Designing Artificial Intelligence and Data-Driven Technologies},
  author={Hossain, S. and Ahmed, S. I.},
  journal={arXiv preprint arXiv:2104.04072},
  year={2021}
}

@inproceedings{jeong2022fairness,
  title={Fairness without imputation: A decision tree approach for fair prediction with missing values},
  author={Jeong, H. and Wang, H. and Calmon, F. P.},
  booktitle={Proceedings of the AAAI Conference on Artificial Intelligence},
  volume={36},
  number={9},
  pages={9558--9566},
  year={2022}
}

@article{lee2019webuildai,
  title={WeBuildAI: Participatory framework for algorithmic governance},
  author={Lee, M. K. and Kusbit, D. and Kahng, A. and Kim, J. T. and Yuan, X. and Chan, A. and Procaccia, A. D.},
  journal={Proceedings of the ACM on Human-Computer Interaction},
  volume={3},
  number={CSCW},
  pages={1--35},
  year={2019}
}

@inproceedings{birhane2022power,
  title={Power to the people? Opportunities and challenges for participatory AI},
  author={Birhane, Abeba and Isaac, William and Prabhakaran, Vinodkumar and Diaz, Mark and Elish, Madeleine Clare and Gabriel, Iason and Mohamed, Shakir},
  booktitle={Proceedings of the 2nd ACM Conference on Equity and Access in Algorithms, Mechanisms, and Optimization},
  pages={1--8},
  year={2022}
}

@inproceedings{laufer2023optimization,
  title={Optimization’s Neglected Normative Commitments},
  author={Laufer, Benjamin and Gilbert, Thomas and Nissenbaum, Helen},
  booktitle={Proceedings of the 2023 ACM Conference on Fairness, Accountability, and Transparency},
  pages={50--63},
  year={2023}
}

@article{chiusi2020automating,
  title={Automating society report 2020},
  author={Chiusi, Fabio and Alfter, Brigitte and Ruckenstein, Minna and Lehtiniemi, Tuukka},
  year={2020},
  publisher={AlgorithmWatch \& Bertelsmann Stiftung}
}

@article{lepri2018fair,
  title={Fair, transparent, and accountable algorithmic decision-making processes: The premise, the proposed solutions, and the open challenges},
  author={Lepri, Bruno and Oliver, Nuria and Letouz{\'e}, Emmanuel and Pentland, Alex and Vinck, Patrick},
  journal={Philosophy \& Technology},
  volume={31},
  pages={611--627},
  year={2018},
  publisher={Springer}
}

@article{barocas2016big,
  title={Big data's disparate impact},
  author={Barocas, Solon and Selbst, Andrew D},
  journal={Calif. L. Rev.},
  volume={104},
  pages={671},
  year={2016},
  publisher={HeinOnline}
}

@book{floridi2021ethics,
  title={Ethics, governance, and policies in artificial intelligence},
  author={Floridi, Luciano and others},
  year={2021},
  publisher={Springer}
}

@article{council2024regulation,
  title={Regulation (EU) 2024/1689 of the European Parliament and of the Council of 13 June 2024 Laying Down Harmonised Rules on Artificial Intelligence and Amending Regulations (EC) No 300/2008,(EU) No 167/2013,(EU) No 168/2013,(EU) 2018/858,(EU) 2018/1139 and (EU) 2019/2144 and Directives 2014/90/EU,(EU) 2016/797 and (EU) 2020/1828 (Artificial Intelligence Act) Off},
  author={Council, EP},
  journal={J. Eur. Union},
  volume={50},
  pages={202},
  year={2024}
}

@article{hardt2016equality,
  title={Equality of opportunity in supervised learning},
  author={Hardt, Moritz and Price, Eric and Srebro, Nati},
  journal={Advances in neural information processing systems},
  volume={29},
  year={2016}
}

@article{obermeyer2019dissecting,
  title={Dissecting racial bias in an algorithm used to manage the health of populations},
  author={Obermeyer, Ziad and Powers, Brian and Vogeli, Christine and Mullainathan, Sendhil},
  journal={Science},
  volume={366},
  number={6464},
  pages={447--453},
  year={2019},
  publisher={American Association for the Advancement of Science}
}

@inproceedings{petersen2023assessing,
  title={On (assessing) the fairness of risk score models},
  author={Petersen, Eike and Ganz, Melanie and Holm, Sune and Feragen, Aasa},
  booktitle={Proceedings of the 2023 ACM Conference on Fairness, Accountability, and Transparency},
  pages={817--829},
  year={2023}
}

@article{pessach2022review,
  title={A review on fairness in machine learning},
  author={Pessach, Dana and Shmueli, Erez},
  journal={ACM Computing Surveys (CSUR)},
  volume={55},
  number={3},
  pages={1--44},
  year={2022},
  publisher={ACM New York, NY}
}

@article{mehrabi2021survey,
  title={A survey on bias and fairness in machine learning},
  author={Mehrabi, Ninareh and Morstatter, Fred and Saxena, Nripsuta and Lerman, Kristina and Galstyan, Aram},
  journal={ACM computing surveys (CSUR)},
  volume={54},
  number={6},
  pages={1--35},
  year={2021},
  publisher={ACM New York, NY, USA}
}

@inproceedings{chan2024group,
  title={Group Fairness via Group Consensus},
  author={Chan, Eunice and Liu, Zhining and Qiu, Ruizhong and Zhang, Yuheng and Maciejewski, Ross and Tong, Hanghang},
  booktitle={The 2024 ACM Conference on Fairness, Accountability, and Transparency},
  pages={1788--1808},
  year={2024}
}

@article{zhang2023deliberating,
  title={Deliberating with AI: improving decision-making for the future through participatory AI design and stakeholder deliberation},
  author={Zhang, Angie and Walker, Olympia and Nguyen, Kaci and Dai, Jiajun and Chen, Anqing and Lee, Min Kyung},
  journal={Proceedings of the ACM on Human-Computer Interaction},
  volume={7},
  number={CSCW1},
  pages={1--32},
  year={2023},
  publisher={ACM New York, NY, USA}
}

@inproceedings{cachel2024prefair,
  title={PreFAIR: Combining Partial Preferences for Fair Consensus Decision-making},
  author={Cachel, Kathleen and Rundensteiner, Elke},
  booktitle={The 2024 ACM Conference on Fairness, Accountability, and Transparency},
  pages={1133--1149},
  year={2024}
}

@article{araujo2020ai,
  title={In AI we trust? Perceptions about automated decision-making by artificial intelligence},
  author={Araujo, Theo and Helberger, Natali and Kruikemeier, Sanne and De Vreese, Claes H},
  journal={AI \& society},
  volume={35},
  number={3},
  pages={611--623},
  year={2020},
  publisher={Springer}
}

@article{grimmelikhuijsen2023explaining,
  title={Explaining why the computer says no: Algorithmic transparency affects the perceived trustworthiness of automated decision-making},
  author={Grimmelikhuijsen, Stephan},
  journal={Public Administration Review},
  volume={83},
  number={2},
  pages={241--262},
  year={2023},
  publisher={Wiley Online Library}
}

@article{chen2020causalml,
  title={Causalml: Python package for causal machine learning},
  author={Chen, Huigang and Harinen, Totte and Lee, Jeong-Yoon and Yung, Mike and Zhao, Zhenyu},
  journal={arXiv preprint arXiv:2002.11631},
  year={2020}
}

@inproceedings{maas2024beyond,
  title={Beyond Participatory AI},
  author={Maas, Jonne and Ingl{\'e}s, Aar{\'o}n Moreno},
  booktitle={Proceedings of the AAAI/ACM Conference on AI, Ethics, and Society},
  volume={7},
  pages={932--942},
  year={2024}
}

@inproceedings{feffer2023preference,
  title={From preference elicitation to participatory ML: A critical survey \& guidelines for future research},
  author={Feffer, Michael and Skirpan, Michael and Lipton, Zachary and Heidari, Hoda},
  booktitle={Proceedings of the 2023 AAAI/ACM Conference on AI, Ethics, and Society},
  pages={38--48},
  year={2023}
}

@article{hornik1989multilayer,
  title={Multilayer feedforward networks are universal approximators},
  author={Hornik, Kurt and Stinchcombe, Maxwell and White, Halbert},
  journal={Neural networks},
  volume={2},
  number={5},
  pages={359--366},
  year={1989},
  publisher={Elsevier}
}

@article{vapnik1999overview,
  title={An overview of statistical learning theory},
  author={Vapnik, Vladimir N},
  journal={IEEE transactions on neural networks},
  volume={10},
  number={5},
  pages={988--999},
  year={1999},
  publisher={IEEE}
}

@article{kalai1975other,
  title={Other solutions to Nash's bargaining problem},
  author={Kalai, Ehud and Smorodinsky, Meir},
  journal={Econometrica: Journal of the Econometric Society},
  pages={513--518},
  year={1975},
  publisher={JSTOR}
}

@article{nash1950bargaining,
  title={The bargaining problem},
  author={Nash, John F and others},
  journal={Econometrica},
  volume={18},
  number={2},
  pages={155--162},
  year={1950}
}

@article{Kelly1998ProportionalFairness,
  author    = {Kelly, F. P. and Maulloo, A. K. and Tan, D. K. H.},
  title     = {Rate control for communication networks: Shadow prices, proportional fairness and stability},
  journal   = {Journal of the Operational Research Society},
  volume    = {49},
  number    = {3},
  pages     = {237--252},
  year      = {1998},
}

@article{Wald1945Minimax,
  author    = {Wald, Abraham},
  title     = {Statistical Decision Functions Which Minimize the Maximum Risk},
  journal   = {Annals of Mathematics},
  volume    = {46},
  number    = {2},
  pages     = {265--280},
  year      = {1945},
  publisher = {Mathematics Department, Princeton University}
}

@incollection{Zeleny1973Compromise,
  author    = {Zeleny, Milan},
  title     = {Compromise Programming},
  booktitle = {Multiple Criteria Decision Making},
  editor    = {Cochrane, J.L. and Zeleny, M.},
  year      = {1973},
  publisher = {University of South Carolina Press},
  address   = {Columbia},
  pages     = {262--301}
}

@book{mohri2018foundations,
  title={Foundations of machine learning},
  author={Mohri, Mehryar and Rostamizadeh, Afshin and Talwalkar, Ameet},
  year={2018},
  publisher={MIT press}
}

@inproceedings{arif2022towards,
  title={Towards substantive conceptions of algorithmic fairness: Normative guidance from equal opportunity doctrines},
  author={Arif Khan, Falaah and Manis, Eleni and Stoyanovich, Julia},
  booktitle={Proceedings of the 2nd ACM Conference on Equity and Access in Algorithms, Mechanisms, and Optimization},
  pages={1--10},
  year={2022}
}

@article{kleinberg2016inherent,
  title={Inherent trade-offs in the fair determination of risk scores},
  author={Kleinberg, Jon and Mullainathan, Sendhil and Raghavan, Manish},
  journal={arXiv preprint arXiv:1609.05807},
  year={2016}
}

@article{chouldechova2017fair,
  title={Fair prediction with disparate impact: A study of bias in recidivism prediction instruments},
  author={Chouldechova, Alexandra},
  journal={Big data},
  volume={5},
  number={2},
  pages={153--163},
  year={2017},
  publisher={Mary Ann Liebert, Inc. 140 Huguenot Street, 3rd Floor New Rochelle, NY 10801 USA}
}

@article{hsu2022pushing,
  title={Pushing the limits of fairness impossibility: Who's the fairest of them all?},
  author={Hsu, Brian and Mazumder, Rahul and Nandy, Preetam and Basu, Kinjal},
  journal={Advances in Neural Information Processing Systems},
  volume={35},
  pages={32749--32761},
  year={2022}
}

@article{arrow1950difficulty,
  title={A difficulty in the concept of social welfare},
  author={Arrow, Kenneth J},
  journal={Journal of political economy},
  volume={58},
  number={4},
  pages={328--346},
  year={1950},
  publisher={The University of Chicago Press}
}

@book{arrow1964social,
  title={Social choice and individual values},
  author={Arrow, Kenneth Joseph and others},
  volume={2},
  year={1964},
  publisher={Wiley New York}
}

@inproceedings{bell2023possibility,
  title={The possibility of fairness: Revisiting the impossibility theorem in practice},
  author={Bell, Andrew and Bynum, Lucius and Drushchak, Nazarii and Zakharchenko, Tetiana and Rosenblatt, Lucas and Stoyanovich, Julia},
  booktitle={Proceedings of the 2023 ACM Conference on Fairness, Accountability, and Transparency},
  pages={400--422},
  year={2023}
}

@article{green2022escaping,
  title={Escaping the impossibility of fairness: From formal to substantive algorithmic fairness},
  author={Green, Ben},
  journal={Philosophy \& Technology},
  volume={35},
  number={4},
  pages={90},
  year={2022},
  publisher={Springer}
}

@article{gibbard1973manipulation,
  title={Manipulation of voting schemes: a general result},
  author={Gibbard, Allan},
  journal={Econometrica: journal of the Econometric Society},
  pages={587--601},
  year={1973},
  publisher={JSTOR}
}

@article{satterthwaite1975strategy,
  title={Strategy-proofness and Arrow's conditions: Existence and correspondence theorems for voting procedures and social welfare functions},
  author={Satterthwaite, Mark Allen},
  journal={Journal of economic theory},
  volume={10},
  number={2},
  pages={187--217},
  year={1975},
  publisher={Elsevier}
}

@book{brandt2016handbook,
  title={Handbook of computational social choice},
  author={Brandt, Felix and Conitzer, Vincent and Endriss, Ulle and Lang, J{\'e}r{\^o}me and Procaccia, Ariel D},
  year={2016},
  publisher={Cambridge University Press}
}

@article{ruadulescu2020multi,
  title={Multi-objective multi-agent decision making: a utility-based analysis and survey},
  author={R{\u{a}}dulescu, Roxana and Mannion, Patrick and Roijers, Diederik M and Now{\'e}, Ann},
  journal={Autonomous Agents and Multi-Agent Systems},
  volume={34},
  number={1},
  pages={10},
  year={2020},
  publisher={Springer}
}

@article{boix2022systematic,
  title={A systematic review on MIVES: A sustainability-oriented multi-criteria decision-making method},
  author={Boix-Cots, David and Pardo-Bosch, Francesc and Blanco, Ana and Aguado, Antonio and Pujadas, Pablo},
  journal={Building and Environment},
  volume={223},
  pages={109515},
  year={2022},
  publisher={Elsevier}
}

@article{chakraborty2023comprehensive,
  title={A comprehensive and systematic review of multi-criteria decision-making methods and applications in healthcare},
  author={Chakraborty, Santonab and Raut, Rakesh D and Rofin, TM and Chakraborty, Shankar},
  journal={Healthcare Analytics},
  volume={4},
  pages={100232},
  year={2023},
  publisher={Elsevier}
}

@article{taherdoost2023multi,
  title={Multi-criteria decision making (MCDM) methods and concepts},
  author={Taherdoost, Hamed and Madanchian, Mitra},
  journal={Encyclopedia},
  volume={3},
  number={1},
  pages={77--87},
  year={2023},
  publisher={MDPI}
}

\clearpage
\section*{Technical Appendix}\label{sec:appendix}

This document provides the extended version of a paper originally presented at the \textit{MODeM@ECAI25} workshop. Compared to the preliminary version, it includes (i) an expanded discussion of the reference literature and (ii) the introduction of analytically derived robustness certificate schemes designed to relax the assumption of non-adversarial stakeholder behavior.

This appendix (i) presents the theoretical proofs for the main results, including complexity bounds and robustness guarantees; (ii) provides detailed descriptions of the compromise functions, actor reward formulations, and case-specific evaluation metrics used in the experiments; (iii) summarizes the orchestration logic of the proposed framework through a unified pseudocode specification covering both training and deployment stages.

Please refer to the \texttt{README.md} file in the available codebase for detailed instructions on setting up the environment and reproducing the experiments.

\subsection*{Proofs}
\subsubsection*{Proof of Theorem 1}

\begin{proof}
We analyze separately the offline (training and selection) and online (inference) phases, focusing only on the additional computational costs relative to a baseline system that performs outcome prediction with cross-validation.

\vspace{0.5em}
\noindent
\textbf{Offline phase (per cross-validation run):}

\begin{enumerate}
    \item \emph{Reward model hyperparameter search.}  
    For each actor $i \in \mathcal{I}$, we perform hyperparameter tuning over $|G_q|$ configurations, each requiring training cost $c_{\mathrm{train}}^q$.  
    Total cost:
    \[
    O\big( |\mathcal{I}| \cdot |G_q| \cdot c_{\mathrm{train}}^q \big).
    \]
    
    \item \emph{Reward model inference on validation set.}  
    For $T_{\mathrm{val}}$ validation points and $|\mathcal{A}|$ actions, we compute reward predictions for all $|\mathcal{I}|$ actors.  
    Total cost:
    \[
    O\big( T_{\mathrm{val}} \cdot |\mathcal{A}| \cdot |\mathcal{I}| \cdot c_{\mathrm{inf}}^q \big).
    \]
    
    \item \emph{Decision strategy evaluations.}  
    For each validation point, we apply $|\mathcal{D}|$ strategies over $|\mathcal{A}|$ actions across all actors adding:
    \[
    O\big( T_{\mathrm{val}} \cdot |\mathcal{A}| \cdot |\mathcal{I}|\cdot |\mathcal{D}| \big).
    \]
    
    \item \emph{Metric evaluation and ranking.}  
    For $|\mathcal{M}|$ evaluation metrics computed across $|\mathcal{D}|$ strategies:
    \[
    O\big( T_{\mathrm{val}} \cdot |\mathcal{D}| \cdot |\mathcal{M}| \big).
    \]
\end{enumerate}

\noindent
Summing these components, the total additional offline cost is:
\[
O\Big( |\mathcal{I}| \cdot (|G_q| \cdot c_{\mathrm{train}}^q + T_{\mathrm{val}} \cdot |\mathcal{A}| \cdot c_{\mathrm{inf}}^q  + T_{\mathrm{val}} \cdot |\mathcal{A}| \cdot |\mathcal{D}|) + T_{\mathrm{val}} \cdot |\mathcal{D}| \cdot |\mathcal{M}| \Big).
\]
\noindent
If we group terms to visually separate train and validation steps, we get: 
\[
O\Big(
|\mathcal{I}| \cdot |G_q| \cdot c_{\mathrm{train}}^q
\;+\; 
T_{\mathrm{val}} \cdot \big[ |\mathcal{I}| \cdot |\mathcal{A}| \cdot (c_{\mathrm{inf}}^q + |\mathcal{D}|) + |\mathcal{D}| \cdot |\mathcal{M}| \big]
\Big)
\]

Or equivalently:
\[
O\Big(
|\mathcal{I}| \cdot \big( |G_q| \cdot c_{\mathrm{train}}^q + T_{\mathrm{val}} \cdot |\mathcal{A}| \cdot ( c_{\mathrm{inf}}^q + |\mathcal{D}| ) \big)
\;+\; T_{\mathrm{val}} \cdot |\mathcal{D}| \cdot |\mathcal{M}|
\Big)
\]

\vspace{0.5em}
\noindent
\noindent\textbf{Online phase (per new deployment context):}

\begin{enumerate}
    \item \emph{Reward inference (actor-specific).}  \\
    For each action $a \in \mathcal{A}$, compute rewards for all $|\mathcal{I}|$ actors:
    \[
    O\big( |\mathcal{A}| \cdot |\mathcal{I}| \cdot c_{\mathrm{inf}}^q \big).
    \]

    \item \emph{Decision strategy application.}  \\
    For each action $a \in \mathcal{A}$, aggregate $|\mathcal{I}|$ rewards (e.g., sum or weighted average) to compute $D^* \big(\{\mathbb{E}[q_i(a \mid \boldsymbol{x'})]\}\big)$:
    \[
    O\big( |\mathcal{A}| \cdot |\mathcal{I}| \big).
    \]
    If evaluating all $D \in \mathcal{D}$, repeat for each strategy:
    \[
    O\big( |\mathcal{A}| \cdot |\mathcal{I}| \cdot |\mathcal{D}| \big).
    \]
\end{enumerate}

\noindent
Total additional online overhead:
\[
\begin{aligned}
O\Big(& |\mathcal{A}| \cdot |\mathcal{I}| \cdot (c_{\mathrm{inf}}^q + 1) \Big) & \text{(best } D^*), \\
O\Big(& |\mathcal{A}| \cdot |\mathcal{I}| \cdot (c_{\mathrm{inf}}^q + |\mathcal{D}|) \Big) & \text{(all } D \in \mathcal{D}).
\end{aligned}
\]

\end{proof}

\subsubsection*{Proof of Theorem 2}
\begin{proof}
\textbf{Step 1 (First-order expansion).}
For any $\Delta\in\mathcal U_\infty(\delta)$,
\[
  \mathcal S_j^{\tau}(\mathbf E+\Delta)
  \;=\;
  \mathcal S_j^{\tau}(\mathbf E)
  +\langle\nabla\mathcal S_j^{\tau}(\mathbf E),\Delta\rangle
  +R_2(\Delta),
\]
where $R_2$ collects second-order terms.

\textbf{Step 2 (Bounding the linear part).}
Because $\Delta_{x,i,a}=0$ for $i\notin V$
and $\|\Delta\|_\infty\le\delta$, Hölder’s inequality and the choice
$\Delta_{x,i,a}=-\delta\,\mathrm{sign}\!\big((\nabla\mathcal S_j^\tau)_{x,i,a}\big)$
for $i\in V$ give
\[
  \langle\nabla\mathcal S_j^{\tau}(\mathbf E),\Delta\rangle
  \;\ge\;
  -\,\delta\,
  \bigl\|\nabla_V\mathcal S_j^{\tau}(\mathbf E)\bigr\|_1.
\]

\textbf{Step 3 (Bounding the quadratic remainder).}
By $\ell_\infty$–smoothness with constant 
$L=\beta_{C_j}(v)/(\tau^{2}\mu^{2})$ and the inner tube,
\[
  |R_2(\Delta)|\le
  \frac{L}{2}\|\Delta\|_\infty^{2}
  \le
  \frac{\delta^{2}\beta_{C_j}(v)}{2\tau^{2}\mu^{2}}.
\]

\textbf{Step 4 (Bias transfer to the sharp score).}
By Assumption~\ref{ass:bias},
\(
  \mathcal S_j^{0}(\mathbf E+\Delta)
  \ge
  \mathcal S_j^{\tau}(\mathbf E+\Delta)-\mathcal B_v^\tau,
\)
with $\mathcal B_v^\tau\le \kappa_{C_j}(v)\tau$.

\textbf{Step 5 (Combine and take the infimum).}
Combining Steps 1--4,
\[
\mathcal S_j^{0}(\mathbf E+\Delta)
\;\ge\;
\mathcal S_j^\tau(\mathbf E)
-\delta\|\nabla_V\mathcal S_j^\tau(\mathbf E)\|_1
-\frac{\delta^{2}\beta_{C_j}(v)}{2\tau^{2}\mu^{2}}
-\mathcal B_v^\tau.
\]
Taking the infimum over $\Delta\in\mathcal U_\infty(\delta)$ yields
the claimed $\mathsf{RLB}_v^\tau(\mathbf E;C_j)$.
\end{proof}

\subsubsection*{Proof of Lemma 3}
\begin{proof}
Define
\(
  g(\tau):=\frac{\delta^{2}\beta_{C_j}(v)}{2\tau^{2}\mu^{2}}
           +\kappa_{C_j}(v)\tau.
\)
\[
  g'(\tau)=
      -\frac{\delta^{2}\beta_{C_j}(v)}{\tau^{3}\mu^{2}}
      +\kappa_{C_j}(v),
  \quad
  g''(\tau)=\frac{3\delta^{2}\beta_{C_j}(v)}{\tau^{4}\mu^{2}}>0.
\]

Setting $g'(\tau)=0$ yields
\(
  \tau^{*3}=
  \frac{\delta^{2}\beta_{C_j}(v)}{\kappa_{C_j}(v)\mu^{2}},
\)
hence the stated cube-root expression. Strict convexity
($g''>0$) guarantees uniqueness. 
\end{proof}

\subsubsection*{Proof of Corollary 4}
\begin{proof}
Plug $\tau=\tau^*$ from Lemma 3
into the bias bound
$\mathcal B_v^\tau\le\kappa_{C_j}(v)\tau$ and note that
this inequality holds uniformly over
$B_\infty(\mathbf E,\delta)$.
\end{proof}

\subsubsection*{Proof of Lemma 5}
\begin{proof}
Fix any pair $(g,h)$ with $g<h$.  By Corollary 4,
\[
  |S_g^{\tau_g^*}-S_g^{0}|\le\epsilon_g,
  \quad
  |S_h^{\tau_h^*}-S_h^{0}|\le\epsilon_h.
\]
Hence
\[
  \bigl|
    (S_g^{\tau_g^*}-S_h^{\tau_h^*})
    -
    (S_g^{0}-S_h^{0})
  \bigr|
  \le\epsilon_g+\epsilon_h.
\]
If the sharp gap condition holds, the smooth gap
remains positive, so the pairwise order is preserved.
Repeating for every pair yields preservation of the total
order. 
\end{proof}

\subsection*{Further experimental details}

\subsubsection*{Compromise Functions}

Below the list of compromise functions used in the experiments:

\paragraph{Nash Bargaining Solution}  
\[
C_{NBS} = \arg\max_{a \in \mathcal{A}} \prod_{i=1}^{N} \bigl( \mathbb{E}[R_i(a \mid  \boldsymbol{x})] - d_i \bigr),
\]  
this function maximizes the product of utility gains above actor-specific disagreement payoffs, balancing fairness and efficiency.

\paragraph{Proportional Fairness}  
\[
C_{PF} = \arg\max_{a \in \mathcal{A}} \sum_{i=1}^{N} \log\bigl(\mathbb{E}[R_i(a \mid  \boldsymbol{x})]\bigr),
\]  
it promotes balanced improvements in collective well-being, ensuring fair trade-offs across actors.

\paragraph{Nash Social Welfare}  
\[
C_{NSW} = \arg\max_{a \in \mathcal{A}} \prod_{i=1}^{N} \mathbb{E}[R_i(a \mid  \boldsymbol{x})],
\]  
this function maximizes the product of actors’ utilities, which is equivalent to proportional fairness under a logarithmic transformation.

\paragraph{Maximin}  
\[
C_{MM} = \arg\max_{a \in \mathcal{A}} \min_{i=1}^{N} \mathbb{E}[R_i(a \mid \boldsymbol{x})],
\]  
it safeguards the most disadvantaged actor by maximizing the minimum utility across all actors.

\paragraph{Compromise Programming (L2)}   
\[
C_{CP-L2} = \arg\min_{a \in \mathcal{A}} \sqrt{\sum_{i=1}^{N} w_i \bigl( u_i^* - \mathbb{E}[R_i(a \mid \boldsymbol{x})] \bigr)^2},
\]  
this approach minimizes the weighted Euclidean distance between the actors’ achieved utilities and their ideal points \( u_i^* \).

\paragraph{Kalai-Smorodinsky Solution} 
\[
C_{KS} = \arg\max_{a \in \mathcal{A}} \min_{i=1}^{N} \frac{\mathbb{E}[R_i(a \mid  \boldsymbol{x})] - d_i}{u_i^* - d_i},
\]  
it maximizes the proportional gains toward each actor’s ideal payoff relative to their disagreement payoff.

\subsubsection*{Actor Reward Functions}
The precise code used to implement the reward functions is available in the utils/rewards/get\_rewards.py script, within the codebase folder provided with the submission. Below is a more precise narrative description. 

\paragraph{Lending Scenario}

\begin{itemize}
    \item \textbf{Bank}: Seeks to maximize profitability. Rewards are tied to repayment probabilities, assigning higher rewards for fully repaid loans and lower rewards for partial or non-repayment.
    \item \textbf{Applicant}: Prioritizes loan access to meet financial needs. Rewards reflect the utility derived from loan approval, modulated by the obligations of repayment. Full approval typically yields higher rewards, while partial or no approval reduces applicant satisfaction.
    \item \textbf{Regulatory Body}: Responsible for ensuring financial stability and inclusivity in lending practices. Rewards balance the stability of the financial system with the willingness to promote social inclusion and provide access to credit for vulnerable applicants.
\end{itemize}

\paragraph{Healthcare Scenario}

\begin{itemize}
    \item \textbf{Healthcare Provider}: Focused on improving patient outcomes while managing costs. Rewards are based on normalized outcome improvements relative to a baseline, with penalties for higher treatment costs.
    \item \textbf{Policy Maker}: Committed to maximizing societal benefits and promoting fairness. Rewards emphasize outcome improvements normalized by potential gains, with additional weighting to promote equity across demographic groups.
    \item \textbf{Parent}: Prioritizes the well-being of their child. Rewards are directly proportional to normalized outcomes, reflecting the straightforward utility parents derive from improved health or cognitive scores.
\end{itemize}

\subsubsection*{Case-Specific Metrics}
The precise code used to compute the metrics reported in the experiments is available in the scripts located in the utils/metrics directory, within the codebase folder provided with the submission. Below, we provide a narrative description of the case-specific metrics employed.

\paragraph{Lending Scenario}

\begin{itemize}
    \item \textbf{Total Profit}: The fraction of potential profit realized from fully or partially repaid granted loans, normalized by the total potential profit across all loans.
    \item \textbf{Total Loss}: The fraction of potential loss incurred from partially repaid or defaulted granted loans, normalized by the total potential loss.
\end{itemize}

\paragraph{Healthcare Scenario}

\begin{itemize}
    \item \textbf{Percentage Treated}: The fraction of individuals receiving the active intervention.
    \item \textbf{Average Outcome Difference}: The average improvement in outcome between treated and control groups.
    \item \textbf{Total Cognitive Score}: The combined cognitive outcome scores from both treated and control groups.
    \item \textbf{Mean Outcome (Treated / Control)}: The average outcome separately for treated and control groups.
    \item \textbf{Cost Effectiveness}: The outcome improvement per unit cost, comparing the relative improvement in treated groups against the differential cost between treatments.
\end{itemize}
\subsection*{Pseudocode}
We summarize the core stages of the proposed participatory training and decision framework in Algorithm~\ref{alg:general_framework}, detailing its offline learning and online deployment steps.

\begin{algorithm}[h!]
\caption{Participatory Training and Decision Framework}
\label{alg:general_framework}
\textbf{Input:} 
Base dataset $\mathcal{T}$; reward-augmented dataset $\mathcal{T}^+$; 
actor set $\mathcal{I}$; action space $\mathcal{A}$; outcome space $\mathcal{O}$; 
strategy set $\mathcal{D}$; evaluation metrics $\mathcal{M}$.\\
\textbf{Output:} Optimal decision strategy $D^*$ and recommended action $a^*$ for new context $\boldsymbol{x}'$.
\begin{algorithmic}[1]
\STATE \textbf{[Offline: Model Training \& Strategy Selection]}
\STATE Train outcome predictor $f$ on $\mathcal{T}$.
\FOR{each actor $i \in \mathcal{I}$}
  \STATE Train reward model $q_i$ on $\mathcal{T}^+_{\mathrm{sampled}} \subseteq \mathcal{T}^+$.
\ENDFOR
\FOR{each decision strategy $D \in \mathcal{D}$}
  \STATE Using $f$ and $\{q_i\}_{i\in\mathcal{I}}$, compute validation matrices $\{\mathbf{E}(\boldsymbol{x}_t)\}_{t=1}^{T_{\mathrm{val}}}$.
  \STATE Compute composite score $\mathcal{S}(\mathbf{E}; D)$ using metrics $\mathcal{M}$.
\ENDFOR
\STATE $D^* \gets \arg\max_{D \in \mathcal{D}} \mathcal{S}(\mathbf{E}; D)$.
\STATE \textbf{[Online: Action Recommendation]}
\STATE Given new context $\boldsymbol{x}'$, construct $\mathbf{E}(\boldsymbol{x}')$ from $f$ and $\{q_i\}$.
\STATE $a^* \gets D^*(\mathbf{E}(\boldsymbol{x}'); \boldsymbol{p})$.
\RETURN $a^*$.
\end{algorithmic}
\end{algorithm}

\end{document}